\documentclass[conference]{IEEEtran}
\IEEEoverridecommandlockouts

\usepackage{cite}
\usepackage{amsmath,amssymb,amsfonts}
\usepackage{algorithmic}
\usepackage{graphicx}
\usepackage{array}
\usepackage{pifont}
\usepackage{amssymb}
\usepackage{textcomp}
\usepackage{xcolor}
\usepackage{multirow}
\usepackage{url}

\def\BibTeX{{\rm B\kern-.05em{\sc i\kern-.025em b}\kern-.08em
    T\kern-.1667em\lower.7ex\hbox{E}\kern-.125emX}}
\begin{document}

\title{ MHAD: Multimodal Home Activity Dataset with Multi-Angle Videos and Synchronized Physiological Signals
}

\author{
    \IEEEauthorblockN{Lei Yu$^{1,2}$, Jintao Fei$^{1\textsuperscript{*}}$, Xinyi Liu$^1$, Yang Yao$^1$, Jun Zhao$^1$, Guoxin Wang$^{1,3}$, Xin Li$^1$}
    \IEEEauthorblockA{$^1$ JD Health, Beijing, China}
    \IEEEauthorblockA{$^2$ School of Software Engineering, Huazhong University of Science and Technology, Wuhan, China}
    \IEEEauthorblockA{$^3$ College of Biomedical Engineering and Instrument Science, Zhejiang University, Hangzhou, China}
\IEEEauthorblockA{yulei@hust.edu.cn,  \{feijintao3, liuxinyi74, yaoyang15, zhaojun10, lixin\}@jd.com,\ guoxin.wang@zju.edu.cn
}
\thanks{\textsuperscript{*}Corresponding author}

}

\maketitle

\begin{abstract}

Video-based physiology, exemplified by remote photoplethysmography (rPPG), extracts physiological signals such as pulse and respiration by analyzing subtle changes in video recordings. This non-contact, real-time monitoring method holds great potential for home settings. Despite the valuable contributions of public benchmark datasets to this technology, there is currently no dataset specifically designed for passive home monitoring. Existing datasets are often limited to close-up, static, frontal recordings and typically include only 1-2 physiological signals.

To advance video-based physiology in real home settings, we introduce the MHAD dataset. It comprises 1,440 videos from 40 subjects, capturing 6 typical activities from 3 angles in a real home environment. Additionally, 5 physiological signals were recorded, making it a comprehensive video-based physiology dataset. MHAD is compatible with the rPPG-toolbox\cite{rppgtoolbox} and has been validated using several unsupervised and supervised methods. Our dataset is publicly available at \url{https://github.com/jdh-algo/MHAD-Dataset}.
\end{abstract}

\begin{IEEEkeywords}
Video-based Physiology, Remote Photoplethysmography (rPPG), Home Environment, MHAD Dataset, Non-contact Monitoring.
\end{IEEEkeywords}

\section{Introduction}

Video-based physiology, with remote photoplethysmography (rPPG) being the most common technique, extracts photoplethysmographic (PPG) waves, respiration movements, and other physiological signals by detecting subtle changes between frames, such as light reflection from the skin\cite{nature1}. This contactless, comfortable, and convenient method is ideal for health monitoring in home settings. It tackles the high costs and complex operations associated with traditional medical devices, which often make regular home monitoring impractical\cite{kumar2015distanceppg}\cite{kranjec2014non}.

Despite its advantages, its video-based measurement performance is highly sensitive to activities (e.g., body movements) and camera angles (e.g., side views). Motion artifacts in the video pose significant challenges, and current state-of-the-art models struggle to generate accurate waveform during activities\cite{wijshoff2016reduction}. Variations in facial angles alter light reflection characteristics and obscure regions with the highest blood perfusion, complicating measurements and reducing accuracy.

\begin{table*}[htbp]
\caption{Comparative Analysis of Datasets }
    \centering
    \resizebox{\textwidth}{!}{
    \begingroup
    \fontsize{11}{14}\selectfont 
    \begin{tabular}{>{\centering\arraybackslash}m{3.0cm} 
                    >{\centering\arraybackslash}m{1cm} 
                    >{\centering\arraybackslash}m{1cm} 
                    >{\centering\arraybackslash}m{4cm}
                    >{\centering\arraybackslash}m{2cm} 
                    >{\centering\arraybackslash}m{4cm} 
                    >{\centering\arraybackslash}m{1cm} 
                    >{\centering\arraybackslash}m{2.5cm} 
                    >{\centering\arraybackslash}m{1cm} 
                    >{\centering\arraybackslash}m{1.3cm}       
                    >{\centering\arraybackslash}m{1.3cm}}
        \hline
        Dataset & Videos & Subjects & Camera & Resolution & Sensors & Signal Types & Non-close-up Shooting & Multi-angle & Exercise & Activities \\
        \hline
        UBFC-PRPG \cite{UBFC} & 42 & 42 & Logitech C920 HD Pro &  640*480 & CMS50E & 1 &  \ding{56}  & \ding{56} & \ding{56}&\ding{56} \\
        PURE \cite{PURE} & 60 & 10 & eco274CVGE &  640*480 & CMS50E & 2 &   \ding{56}  & \ding{56}  & \ding{56}&\ding{56} \\
        COHFACE \cite{COHFACE} & 160 & 40 & Logitech HD C525 &  640*480 & Thought Technology & 2 &   \ding{56}  & \ding{56}  & \ding{56}&\ding{56} \\
        MMPD \cite{MMPD} & 660 & 33 & Galaxy S22 Ultra  &  320*240 & HKG-07C+ & 1 &   \ding{56}  &  \ding{56} &   \ding{52}&\ding{52} \\
        \hline
        MHAD(Ours) & 1440 & 40 & WN-L1812.K56R &  1280*720 & BIOPAC MP160 + OMRON J760 & 5 &   \ding{52} &  \ding{52} &  \ding{52}&\ding{52} \\
        \hline
    \end{tabular}
    \endgroup
    }
\end{table*}

The collection and analysis of multimodal datasets have become increasingly important in physiological signal research, spurring advancements in video-based physiology\cite{ref1}\cite{nature3}\cite{zhang2023recent}. However, existing datasets have significant limitations. They typically include frontal video recordings closely aligned with the face, failing to capture natural scenarios from multiple angles. Additionally, these datasets mostly consist of static state videos, with a scarcity of data in active conditions\cite{MMPD}\cite{selvaraju2022continuous}. More importantly, research has shown that videos can be used to extract various physiological signals beyond pulse, such as respiratory rate (RR), oxygen saturation level (SpO2), and blood pressure\cite{nature2}. However, the absence of these physiological signals in existing datasets means that related methods have not been rigorously evaluated. Most datasets only include PPG or heart rate data, lacking other crucial physiological signals, which makes remote methods less accurate in measuring these signals\cite{ref3}.

To address these issues, we introduce the Multimodal Home Activity Dataset (MHAD-Dataset). This dataset is the first to offer multi-angle (frontal, 45-degree side, 90-degree side) synchronized video recordings along with multiple physiological signals in a real home environment. Our dataset encompasses six typical home activities (watching TV, using a phone, reading a book, talking, eating, drinking), ensuring representative data. The recorded physiological signals (respiration, PPG, ECG, SpO2, blood pressure) span a comprehensive range of measurements, providing the most complete set of physiological signals among existing datasets. All physiological signals were collected using gold-standard BIOPAC sensors, and the videos were recorded using commercial USB cameras. Our contributions are as follows:

\begin{itemize}

\item[1)]We introduce the MHAD Dataset, the first public video-based physiological dataset of subjects in a real home environment. It features different camera angles, various household activities, and rich physiological signals, aiming to enhance the accuracy and realism of video-based physiological research.
\item[2)]We conducted a comprehensive quantitative analysis to explore the robustness of current state-of-the-art supervised and unsupervised methods under different camera angles and various activities in our dataset.
\end{itemize}

\section{Related Work}
\label{sec:format}

A significant challenge in advancing video-based physiology is the scarcity of publicly available datasets recorded in real-world conditions. Commonly used datasets like MAHNOB-HCI\cite{HCI} and DEAP\cite{deap}, initially designed for emotion recognition, involve lighting changes and movements related to eliciting emotional responses, making them suboptimal for evaluating rPPG algorithms in complex real-world scenarios. There are some datasets specifically designed for rPPG, such as UBFC-RPPG\cite{UBFC}, PURE\cite{PURE}, COHFACE\cite{COHFACE}, and MMPD\cite{MMPD}. We chose to analyze and compare these datasets.

The UBFC-RPPG\cite{UBFC} dataset, captured with a Logitech C920 HD Pro webcam at 640x480 resolution and 30fps, includes ground truth PPG waveform from a CMS50E pulse oximeter. Subjects sat approximately 1 meter from the camera. While reliable and widely used, its single angle, limited body movement, and lack of real-world scenarios restrict its applicability. Moreover, it only provides heart rate labels.

The PURE\cite{PURE} dataset features 60 one-minute sequences captured with an eco274CVGE camera at 30fps (640x480 resolution), with PPG data collected at 60Hz. It includes various movements such as talking and head rotation but lacks diversity in real-world tasks and multi-angle shooting. Additionally, it does not cover a comprehensive range of physiological signals.

The COHFACE\cite{COHFACE} dataset, recorded using a Logitech HD C525 at 640x480 resolution and 20fps, includes PPG and respiration data collected with Thought Technology equipment. Despite its widespread use, its single camera angle and close-up face shots limit its real-world applicability, and it only collects two types of physiological signals: blood volume pulse and breathing rate.

The MMPD\cite{MMPD} dataset, recorded with a Samsung Galaxy S22 Ultra at 320x240 resolution, includes 660 one-minute videos and PPG signals from an HKG-07C+ oximeter. The videos cover four activities: stillness, head rotation, talking, and taking selfies. However, the subjects recorded with phones close to their faces, which does not meet the needs for contactless monitoring in home environments. Additionally, the limited variety of activities fails to encompass most home tasks, and the dataset only provides ground truth for ppg signals.

\section{DATASET}
\label{sec:pagestyle}

To create a dataset that captures the diversity and complexity of activities in real home environments, we recruited 40 voluntary participants including 10 women and 30 men aged 23-50, and conducted recordings in actual living rooms. To minimize the impact of lighting changes, we used fixed, moderately intense light sources during recording sessions. Three USB cameras were used to synchronously record videos, resulting in a total of 1,440 videos, each 30 seconds long. Concurrently, we used BIOPAC MP160 and OMRON J760 to record 5 raw signals (respiration, PPG, ECG, SpO2, blood pressure) and 2 calculated signals (heart rate, respiratory rate), all millisecond-level synchronized with the videos.


\begin{figure*}[htbp]
    \centering
    \includegraphics[width=0.95\textwidth]{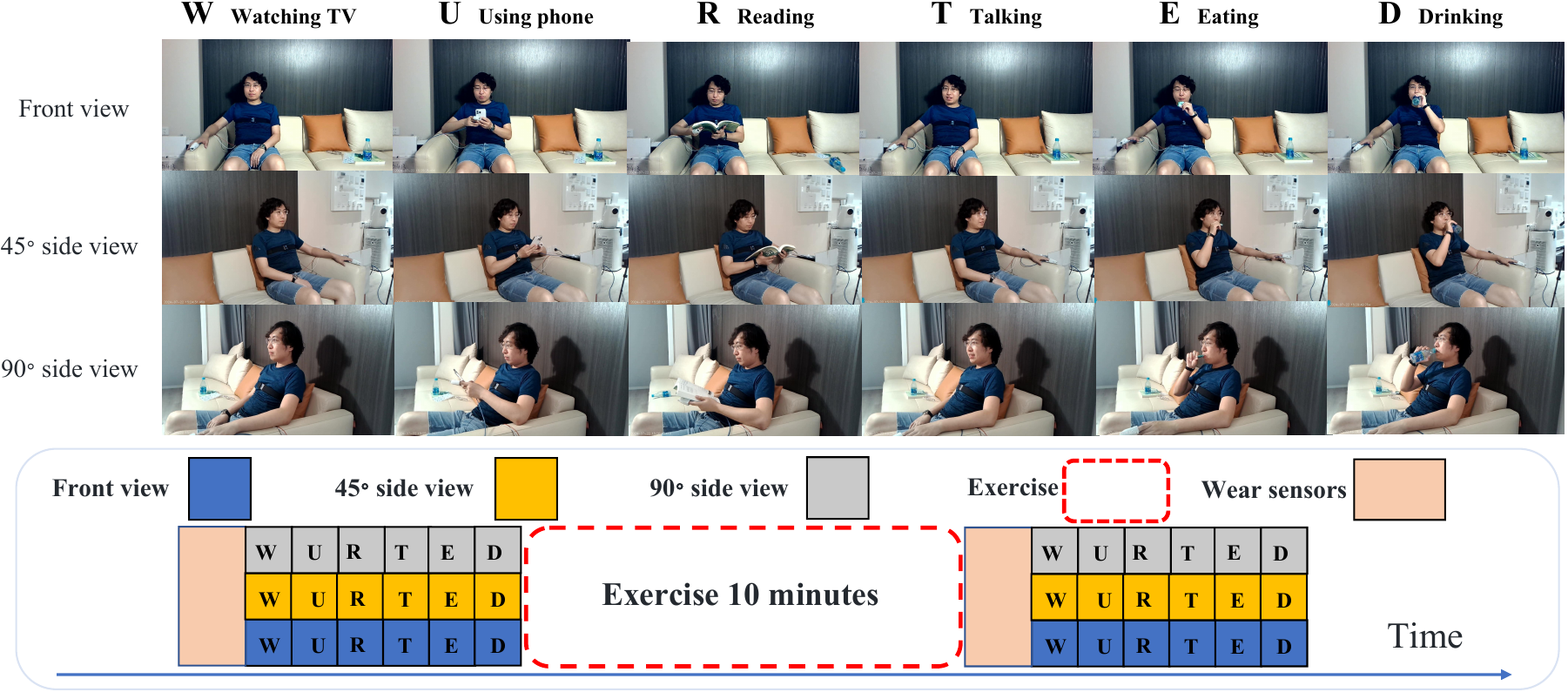} 
    \caption{Sample Frames and Data Collection Pipeline of the MHAD Dataset}
    \label{fig:example_image}
     \vspace{-0.5em}  
\end{figure*}

\subsection{Data Collecting}

As previously mentioned, camera angles and motion significantly impact the extraction of physiological signals from video. In MHAD, cameras recorded from three different angles (frontal, 45-degree side, and 90-degree side). The data collection pipeline is illustrated in Figure \ref{fig:example_image}. To better reflect home environments, all videos were captured from a distance of 2 meters, unlike the close-up recordings in other datasets.
For the activities, we designed tasks representing six common household scenarios: watching TV, using phone, reading, talking, eating, and drinking.  After completing stage 1, subjects were asked to run or climb stairs for about ten minutes to accelerate their pulse and respiratory rate. Stage 2 was then recorded in the same manner as stage 1. Blood pressure measurements were conducted within 1 minute before each stage. To ensure millisecond-level alignment, we recorded timestamps for the videos and ground truth physiological signals and then aligned them.

\subsection{Data Processing and Storage}

Videos were 30 seconds at 30 fps, 1280x720 pixels. Figure 2 shows some sample frames from MHAD. Respiration, PPG, ECG, and SpO2 were recorded at 2000 Hz and filtered using the default filters provided by BIOPAC. Heart rate, respiratory rate, systolic pressure, and diastolic pressure were recorded as either mean values or single measurements. All signals were saved in a .csv file for each activity, corresponding to the videos from three angles.

\section{EXPERIMENTS}
\label{sec:typestyle}

We evaluated mainstream remote heart rate and respiratory rate estimation methods on MHAD, including both supervised neural network methods and unsupervised methods. We also hope that the release of MHAD will encourage advancements in contactless measurement for other physiological signals.

\subsection{Unsupervised Signal Processing Methods}

\begin{table*}[htbp]
\centering
\caption{Benchmarking Heart Rate and Respiratory Rate Measurements Using Various Unsupervised Methods on MHAD.}
\resizebox{0.76\textwidth}{!}{%
\begin{tabular}{l|cc|cc|cc|cc}
\hline
\multirow{3}{*}{Method} & \multicolumn{6}{c|}{Heart rate} & \multicolumn{2}{c}{Respiratory rate} \\
\cline{2-9}
 & \multicolumn{2}{c|}{ICA\cite{ICA}} & \multicolumn{2}{c|}{POS\cite{POS}} & \multicolumn{2}{c|}{PBV\cite{PBV}} & \multicolumn{2}{c}{Tan et al.\cite{lightweight}} \\
\cline{2-9}
 & MAE $\downarrow$ & RMSE $\downarrow$ & MAE $\downarrow$ & RMSE $\downarrow$ & MAE $\downarrow$ & RMSE $\downarrow$ & MAE $\downarrow$ & RMSE $\downarrow$ \\
\hline
\multicolumn{9}{l}{Shooting Angle} \\
\hline
Fontal & 14.55 & 20.90 &\textbf{10.98} & \textbf{19.87} & 16.56 & 24.18 & 4.39 & 6.07 \\
45-degree side & 28.88 & 39.24 & 41.96 & 52.10 & 34.87 & 44.72 & 4.22 & 5.89 \\
90-degree side & 26.19 & 35.20 & 49.54 & 57.20 & 39.20 & 48.22 & \textbf{3.95} & \textbf{5.54} \\
\hline
\multicolumn{9}{l}{Activity(Fontal)} \\
\hline
Watching TV & 7.18 & \textbf{12.58} & \textbf{7.00} & 17.12 & 9.47 & 17.17 & 3.10 & \textbf{4.51} \\
Using phone & 12.63 & 19.64 & 9.73 & 18.37 & 15.62 & 24.97 & \textbf{3.05} & 4.66 \\
Reading & 12.83 & 20.62 & 15.95 & 27.66 & 17.27 & 25.90 & 3.60 & 4.83 \\
Talking & 18.91 & 25.09 & 9.31 & 18.22 & 16.50 & 23.88 & 4.35 & 5.89 \\
Eating & 13.27 & 17.85 & 10.85 & 17.58 & 17.53 & 24.02 & 5.80 & 7.84 \\
Drinking & 22.47 & 26.53 & 13.02 & 18.27 & 22.98 & 27.77 & 6.45 & 7.71 \\
\hline
\end{tabular}%
}
\label{tab:comparison}
\end{table*}

For unsupervised remote heart rate measurement, we evaluated three traditional unsupervised learning methods on our dataset: ICA\cite{ICA}, POS\cite{POS}, and PBV\cite{PBV}. When comparing shooting angles, we calculated results for all activities across three different angles to isolate the influence of activities. Similarly, in activity comparison experiments, we focused on front-facing videos to exclude angle conditions, thereby isolating the challenges posed by each specific task.

As shown in Table \ref{tab:comparison}, the POS method\cite{POS} performed best on simple front-facing data, but the mean absolute error (MAE) remained above 10 due to the complexity of the activities. The ICA method\cite{ICA} showed relatively better performance on data captured from 45-degree and 90-degree angles, yet it remained highly inaccurate, with an average MAE exceeding 25. Although side-angle videos are more realistic, the accuracy of unsupervised methods remains low, rendering them nearly unusable.

Regarding activities, all methods performed well when watching TV, as participants' bodies and heads remained almost motionless, and avoiding facial coverage from lowering their heads (e.g., reading or using phone). Most of other activities showed an MAE greater than 10. Overall, the POS method provided relatively accurate measurements across various activities.

For unsupervised remote respiration measurement, we evaluated the optical flow method\cite{lightweight} on our dataset. As shown in Table \ref{tab:comparison}, for videos from all three angles, the MAE was less than 5, with the 90-degree side angle yielding the best results and the front-facing angle the worst. This is because the side view clearly captures the subtle movements of the chest and abdomen during breathing. For the six activities, the MAE and RMSE were optimal for minimal movement scenarios, such as using the phone and watching TV. However, for activities with larger movements, the errors increased, with the MAE for eating and drinking both exceeding 5.

\subsection{Supervised Deep Learning Methods}

\begin{table*}[h!]
\centering
\caption{Benchmarking Heart Rate and Respiratory Rate Measurements Using Different Deep Learning Methods on MHAD.}
\resizebox{0.87\textwidth}{!}{%
\setlength{\tabcolsep}{4pt} 
\begin{tabular}{c|>{\centering\arraybackslash}p{0.7cm}c|>{\centering\arraybackslash}p{0.7cm}c|>{\centering\arraybackslash}p{0.7cm}c|>{\centering\arraybackslash}p{0.7cm}c|>{\centering\arraybackslash}p{0.7cm}c|>{\centering\arraybackslash}p{0.7cm}c}
\hline
 & \multicolumn{6}{c|}{Heart rate} & \multicolumn{6}{c}{Respiratory rate}\\ \hline
Training Set & \multicolumn{6}{c|}{UBFC-RPPG\cite{UBFC}} & \multicolumn{6}{c}{COHFACE\cite{COHFACE}} \\ \hline
Testing Set & \multicolumn{6}{c|}{MHAD} & \multicolumn{6}{c}{MHAD} \\ \hline
Method & \multicolumn{2}{c|}{EfficientPhys\cite{Efficientphys}} & \multicolumn{2}{c|}{TSCAN\cite{TSCAN}} & \multicolumn{2}{c|}{DeepPhys\cite{Deepphys}} & \multicolumn{2}{c|}{EfficientPhys\cite{Efficientphys}} & \multicolumn{2}{c|}{TSCAN\cite{TSCAN}} & \multicolumn{2}{c}{DeepPhys\cite{Deepphys}} \\ \hline
 & {MAE$\downarrow$} & {RMSE$\downarrow$} & {MAE$\downarrow$} & {RMSE$\downarrow$} & {MAE$\downarrow$} & {RMSE$\downarrow$} & {MAE$\downarrow$} & {RMSE$\downarrow$} & {MAE$\downarrow$} & {RMSE$\downarrow$} & {MAE$\downarrow$} & {RMSE$\downarrow$} \\ \hline
 
\multicolumn{13}{l}{Shooting Angle} \\ \hline
Fontal& \textbf{9.30} & \textbf{15.83} & 13.39 & 19.28 & 14.24 & 20.64 & 9.02 & 10.09 & \textbf{8.52} & \textbf{11.58} & 9.10 & 12.15 \\ 
45-degree side & 13.95 & 19.71 & 16.27 & 21.20 & 16.61 & 24.29 & 9.99 & 12.87 & 9.34 & 12.45 & 9.11 & 12.35 \\ 
90-degree side & 17.58 & 22.96 & 17.87 & 22.60 & 19.28 & 26.37 & 9.93 & 12.81 & 9.19 & 12.22 & 9.21 & 12.32 \\ \hline
\multicolumn{13}{l}{Activity(Fontal)} \\ \hline
Watching TV & \textbf{1.97} & \textbf{5.12} & 4.65 & 9.47 & 6.83 & 13.37 & 7.84 & 10.66 & \textbf{6.43} & \textbf{9.58} & 6.56 & 9.99 \\ 
Using phone & 6.13 & 11.99 & 9.66 & 16.10 & 9.82 & 16.46 & 8.26 & 11.59 & 7.51 & 10.81 & 9.20 & 12.03 \\ 
Reading & 10.30 & 16.30 & 14.85 & 20.53 & 15.20 & 21.33 & 7.07 & 9.83 & 8.70 & 11.55 & 9.11 & 11.62 \\ 
Talking & 10.61 & 17.63 & 17.44 & 23.14 & 16.25 & 22.56 & 11.20 & 14.52 & 9.55 & 13.24 & 9.84 & 13.36 \\ 
Eating & 10.26 & 15.62 & 15.13 & 18.86 & 14.98 & 20.11 & 9.16 & 12.34 & 9.53 & 12.18 & 9.73 & 12.71 \\ 
Drinking & 16.32 & 22.46 & 18.63 & 23.85 & 22.39 & 27.15 & 10.61 & 13.04 & 9.42 & 12.12 & 10.15 & 12.87 \\ \hline
\end{tabular}
}
\label{table:comparison}

\end{table*}
 We evaluated the performance of state-of-the-art supervised neural networks for heart rate and respiratory rate measurement using the MHAD dataset, focusing on the impact of camera angles and activities. We employed the EfficientPhys\cite{Efficientphys}, TS-CAN\cite{TSCAN}, and DeepPhys\cite{Deepphys} models, pretrained on the UBFC-RPPG\cite{UBFC} and COHFACE\cite{COHFACE} datasets. The evaluation criteria were consistent with those used for unsupervised methods. Table 3 presents the results across different tasks.

For heart rate, the neural networks demonstrated poor generalizability with side-angle videos, performing well only with frontal recordings. In six activity scenarios, the MAE for frontal recordings exceeded 10 in four scenarios, except for watching TV and using phone. This is mostly due to the training data being predominantly close-up, frontal, and mostly static videos. Among the three methods, EfficientPhys showed the highest accuracy, achieving an MAE of less than 2 in the watching TV scenario and was the only method with an MAE below 10 in frontal angle data.

For respiratory rate, measurement errors were significant across all three angles and six activities. TSCAN performed best while watching TV and with frontal recording. These methods require clear frontal facial captures to measure respiratory rate accurately. Diverse shooting angles, greater recording distances, and varied activity scenarios in our dataset increased the measurement difficulty.

\subsection{Discussion}
The areas with rich blood flow on the face are the forehead, cheeks, nose, and chin. Therefore, when these areas are partially covered (side camera angles or head lowering), the blood flow captured by camera becomes weaker, leading to increased errors in heart rate measurement. Moreover, supervised methods are highly dependent on the similarity between the test data and the training data, so video diversity in camera angles and activities may also contribute to these errors.

For respiratory rate measurement, unsupervised methods significantly outperform deep learning approaches on MHAD. Since the respiratory waveform is derived similarly to rPPG, the errors in measuring respiratory rate using these supervised learning methods also stem from poor facial angles and diverse activities in videos.

To enhance the quality and realism of video-based physiological data collection in passive home monitoring, we recommend increasing recordings that reflect realistic activities, rather than the predominantly stationary videos common in most datasets. Close-up frontal recordings should be minimized, maintaining an appropriate distance and capturing multiple angles to improve real-world usability. Additionally, capturing other vital signs, such as blood pressure and SpO2, alongside pulse and respiration, can advance the remote measurement of all physiological signals. Ensuring the temporal alignment of video and physiological signals is also crucial.

Overall, MHAD enables the training and evaluation of video-based physiological models under multiple angles, various movements, and moderate distances. This can facilitate passive video-based physiological signal measurement in home settings, rather than close-up, frontal, active measurements.

\section{CONCLUSION}
\label{sec:majhead}

In this paper, we introduced the MHAD dataset, which comprises 1,440 videos collected in home environments from 40 subjects, featuring videos from three different angles and six common household activities. The MHAD dataset aims to address the gaps and limitations of existing datasets in passive home monitoring, particularly concerning different shooting angles and activities. Additionally, this dataset includes a comprehensive set of physiological signals in video-based physiology, with the goal of advancing contactless measurement beyond just pulse and respiration. The MHAD dataset and our evaluation of various methods represent a significant step forward in enhancing the accuracy and applicability of these technologies, with the potential to improve home health monitoring and other related applications.

\bibliographystyle{IEEEtran}
\bibliography{IEEEabrv,strings}

\begin{thebibliography}{10}
\providecommand{\url}[1]{#1}
\csname url@samestyle\endcsname
\providecommand{\newblock}{\relax}
\providecommand{\bibinfo}[2]{#2}
\providecommand{\BIBentrySTDinterwordspacing}{\spaceskip=0pt\relax}
\providecommand{\BIBentryALTinterwordstretchfactor}{4}
\providecommand{\BIBentryALTinterwordspacing}{\spaceskip=\fontdimen2\font plus
\BIBentryALTinterwordstretchfactor\fontdimen3\font minus \fontdimen4\font\relax}
\providecommand{\BIBforeignlanguage}[2]{{%
\expandafter\ifx\csname l@#1\endcsname\relax
\typeout{** WARNING: IEEEtran.bst: No hyphenation pattern has been}%
\typeout{** loaded for the language `#1'. Using the pattern for}%
\typeout{** the default language instead.}%
\else
\language=\csname l@#1\endcsname
\fi
#2}}
\providecommand{\BIBdecl}{\relax}
\BIBdecl

\bibitem{rppgtoolbox}
X.~Liu, G.~Narayanswamy, A.~Paruchuri, X.~Zhang, J.~Tang, Y.~Zhang, Y.~Wang, S.~Sengupta, S.~Patel, and D.~McDuff, ``rppg-toolbox: Deep remote ppg toolbox,'' \emph{arXiv preprint arXiv:2210.00716}, 2022.

\bibitem{nature1}
R.~Amelard, C.~Scharfenberger, F.~Kazemzadeh, K.~J. Pfisterer, B.~S. Lin, D.~A. Clausi, and A.~Wong, ``Feasibility of long-distance heart rate monitoring using transmittance photoplethysmographic imaging (ppgi),'' \emph{Scientific reports}, vol.~5, no.~1, p. 14637, 2015.

\bibitem{kumar2015distanceppg}
M.~Kumar, A.~Veeraraghavan, and A.~Sabharwal, ``Distanceppg: Robust non-contact vital signs monitoring using a camera,'' \emph{Biomedical optics express}, vol.~6, no.~5, pp. 1565--1588, 2015.

\bibitem{kranjec2014non}
J.~Kranjec, S.~Begu{\v{s}}, G.~Ger{\v{s}}ak, and J.~Drnov{\v{s}}ek, ``Non-contact heart rate and heart rate variability measurements: A review,'' \emph{Biomedical signal processing and control}, vol.~13, pp. 102--112, 2014.

\bibitem{wijshoff2016reduction}
R.~W. Wijshoff, M.~Mischi, and R.~M. Aarts, ``Reduction of periodic motion artifacts in photoplethysmography,'' \emph{IEEE Transactions on Biomedical Engineering}, vol.~64, no.~1, pp. 196--207, 2016.

\bibitem{UBFC}
S.~Bobbia, R.~Macwan, Y.~Benezeth, A.~Mansouri, and J.~Dubois, ``Unsupervised skin tissue segmentation for remote photoplethysmography,'' \emph{Pattern Recognition Letters}, vol. 124, pp. 82--90, 2019.

\bibitem{PURE}
R.~Stricker, S.~M{\"u}ller, and H.-M. Gross, ``Non-contact video-based pulse rate measurement on a mobile service robot,'' in \emph{The 23rd IEEE International Symposium on Robot and Human Interactive Communication}.\hskip 1em plus 0.5em minus 0.4em\relax IEEE, 2014, pp. 1056--1062.

\bibitem{COHFACE}
G.~Heusch, A.~Anjos, and S.~Marcel, ``A reproducible study on remote heart rate measurement,'' \emph{arXiv preprint arXiv:1709.00962}, 2017.

\bibitem{MMPD}
J.~Tang, K.~Chen, Y.~Wang, Y.~Shi, S.~Patel, D.~McDuff, and X.~Liu, ``Mmpd: multi-domain mobile video physiology dataset,'' in \emph{2023 45th Annual International Conference of the IEEE Engineering in Medicine \& Biology Society (EMBC)}.\hskip 1em plus 0.5em minus 0.4em\relax IEEE, 2023, pp. 1--5.

\bibitem{ref1}
R.~Macwan, Y.~Benezeth, and A.~Mansouri, ``Remote photoplethysmography with constrained ica using periodicity and chrominance constraints,'' \emph{Biomedical engineering online}, vol.~17, pp. 1--22, 2018.

\bibitem{nature3}
X.~Tao, D.~Gao, W.~Zhang, T.~Liu, B.~Du, S.~Zhang, and Y.~Qin, ``A multimodal physiological dataset for driving behaviour analysis,'' \emph{Scientific data}, vol.~11, no.~1, p. 378, 2024.

\bibitem{zhang2023recent}
X.~Zhang, M.~Hu, Y.~Zhang, G.~Zhai, and X.-P. Zhang, ``Recent progress of optical imaging approaches for noncontact physiological signal measurement: A review,'' \emph{Advanced Intelligent Systems}, vol.~5, no.~9, p. 2200345, 2023.

\bibitem{selvaraju2022continuous}
V.~Selvaraju, N.~Spicher, J.~Wang, N.~Ganapathy, J.~M. Warnecke, S.~Leonhardt, R.~Swaminathan, and T.~M. Deserno, ``Continuous monitoring of vital signs using cameras: A systematic review,'' \emph{Sensors}, vol.~22, no.~11, p. 4097, 2022.

\bibitem{nature2}
Y.~Liang, Z.~Chen, G.~Liu, and M.~Elgendi, ``A new, short-recorded photoplethysmogram dataset for blood pressure monitoring in china,'' \emph{Scientific data}, vol.~5, no.~1, pp. 1--7, 2018.

\bibitem{ref3}
M.~A. Almarshad, M.~S. Islam, S.~Al-Ahmadi, and A.~S. BaHammam, ``Diagnostic features and potential applications of ppg signal in healthcare: A systematic review,'' in \emph{Healthcare}, vol.~10, no.~3.\hskip 1em plus 0.5em minus 0.4em\relax MDPI, 2022, p. 547.

\bibitem{HCI}
M.~Soleymani, J.~Lichtenauer, T.~Pun, and M.~Pantic, ``A multimodal database for affect recognition and implicit tagging,'' \emph{IEEE Transactions on Affective Computing}, vol.~3, no.~1, pp. 42--55, 2012.

\bibitem{deap}
S.~Koelstra, C.~Muhl, M.~Soleymani, J.-S. Lee, A.~Yazdani, T.~Ebrahimi, T.~Pun, A.~Nijholt, and I.~Patras, ``Deap: A database for emotion analysis ;using physiological signals,'' \emph{IEEE Transactions on Affective Computing}, vol.~3, no.~1, pp. 18--31, 2012.

\bibitem{ICA}
M.-Z. Poh, D.~J. McDuff, and R.~W. Picard, ``Advancements in noncontact, multiparameter physiological measurements using a webcam,'' \emph{IEEE transactions on biomedical engineering}, vol.~58, no.~1, pp. 7--11, 2010.

\bibitem{POS}
W.~Wang, A.~C. Den~Brinker, S.~Stuijk, and G.~De~Haan, ``Algorithmic principles of remote ppg,'' \emph{IEEE Transactions on Biomedical Engineering}, vol.~64, no.~7, pp. 1479--1491, 2016.

\bibitem{PBV}
G.~De~Haan and A.~Van~Leest, ``Improved motion robustness of remote-ppg by using the blood volume pulse signature,'' \emph{Physiological measurement}, vol.~35, no.~9, p. 1913, 2014.

\bibitem{lightweight}
X.~Tan, M.~Hu, G.~Zhai, Y.~Zhu, W.~Li, and X.-P. Zhang, ``Lightweight video-based respiration rate detection algorithm: An application case on intensive care,'' \emph{IEEE Transactions on Multimedia}, vol.~26, pp. 1761--1775, 2023.

\bibitem{Efficientphys}
X.~Liu, B.~Hill, Z.~Jiang, S.~Patel, and D.~McDuff, ``Efficientphys: Enabling simple, fast and accurate camera-based cardiac measurement,'' in \emph{Proceedings of the IEEE/CVF winter conference on applications of computer vision}, 2023, pp. 5008--5017.

\bibitem{TSCAN}
X.~Liu, J.~Fromm, S.~Patel, and D.~McDuff, ``Multi-task temporal shift attention networks for on-device contactless vitals measurement,'' \emph{Advances in Neural Information Processing Systems}, vol.~33, pp. 19\,400--19\,411, 2020.

\bibitem{Deepphys}
W.~Chen and D.~McDuff, ``Deepphys: Video-based physiological measurement using convolutional attention networks,'' in \emph{Proceedings of the european conference on computer vision (ECCV)}, 2018, pp. 349--365.

\end{thebibliography}


\end{document}